\documentclass[pdflatex,sn-basic,Numbered]{sn-jnl}

\usepackage{graphicx}%
\usepackage{multirow}%
\usepackage{amsmath,amssymb,amsfonts}%
\usepackage{amsthm}%
\usepackage{mathrsfs}%
\usepackage[title]{appendix}%
\usepackage{xcolor}%
\usepackage{textcomp}%
\usepackage{manyfoot}%
\usepackage{booktabs}%

\usepackage{siunitx}%
\usepackage{xspace}%
\usepackage{array}%
\usepackage{mathtools}%
\usepackage{makecell}%

\newcommand{\us}{\_\allowbreak}

\newcounter{mainfig}\newcounter{maintab}

\makeatletter
\DeclareRobustCommand\onedot{\futurelet\@let@token\@onedot}
\def\@onedot{\ifx\@let@token.\else.\null\fi\xspace}

\def\eg{\emph{e.g}\onedot} 
\def\ie{\emph{i.e}\onedot} 
 
\def\etc{\emph{etc}\onedot}

\makeatother

\newcommand{\ms}[2]{\begin{tabular}[c]{@{}c@{}}#1\\[-1pt]{\fontsize{6.5}{7}\selectfont$\pm$#2}\end{tabular}}
\newcommand{\hd}[1]{\begin{tabular}[b]{@{}c@{}}#1\end{tabular}}

\hypersetup{bookmarksdepth=2,
  pdftitle={Continual Humanoid Motion Learning},
  pdfauthor={Zhewen He, Hao Huang, Geeta Chandra Raju Bethala, Anthony Tzes, Yi Fang},
  pdfsubject={Research preprint},
  pdfkeywords={Continual learning, Humanoid robots, Whole-body motion control, Low-rank adaptation, Sim-to-real transfer}}

\begin{document}

\title[Continual Humanoid Motion Learning]{Continual Humanoid Motion Learning}

\author*[1]{\fnm{Zhewen}\sur{He}}\email{zh3510@nyu.edu}
\author[1]{\fnm{Hao}\sur{Huang}}
\author[1]{\fnm{Geeta Chandra Raju}\sur{Bethala}}
\author[1]{\fnm{Anthony}\sur{Tzes}}
\author[1]{\fnm{Yi}\sur{Fang}}

\affil*[1]{\orgname{New York University Abu Dhabi},
  \orgaddress{\city{Abu Dhabi}, \country{United Arab Emirates}}}

\abstract{Humanoid whole-body controllers can now track a diverse set of dynamic motions, but they are typically trained offline and then frozen, so teaching such a controller a new skill tends to erode the skills it already mastered. We study \emph{continual learning for humanoid whole-body motion}, where a single controller must acquire skills from a sequential task stream without revisiting past data. We introduce \emph{Similarity-guided LoRA-PNN}, a progressive neural network (PNN) policy that prevents catastrophic forgetting by construction while reusing knowledge across skills through lightweight low-rank adaptation. A two-level motion-similarity measure, built from dynamic time warping aggregated by optimal transport, decides which prior skill to build on and how much new capacity to allocate, yielding strong forward transfer and large efficiency gains. Across six sequentially learned skill categories, our similarity-guided LoRA policy attains the best forward transfer ($0.125$ vs.\ $0.079$) and the highest average accuracy among all methods, while saving up to $94.5\%$ of trainable parameters and $40.8\%$ of training time. The resulting controller reaches $96.13\%$ sim-to-sim transfer and is deployed on a physical Unitree~G1. Our code is available at \url{https://anonymous.4open.science/r/continual-humanoid-learning-35D3}.}


\maketitle
\begin{center}\small Preprint\end{center}

\begin{figure}[!ht]
\centering
\includegraphics[width=\textwidth]{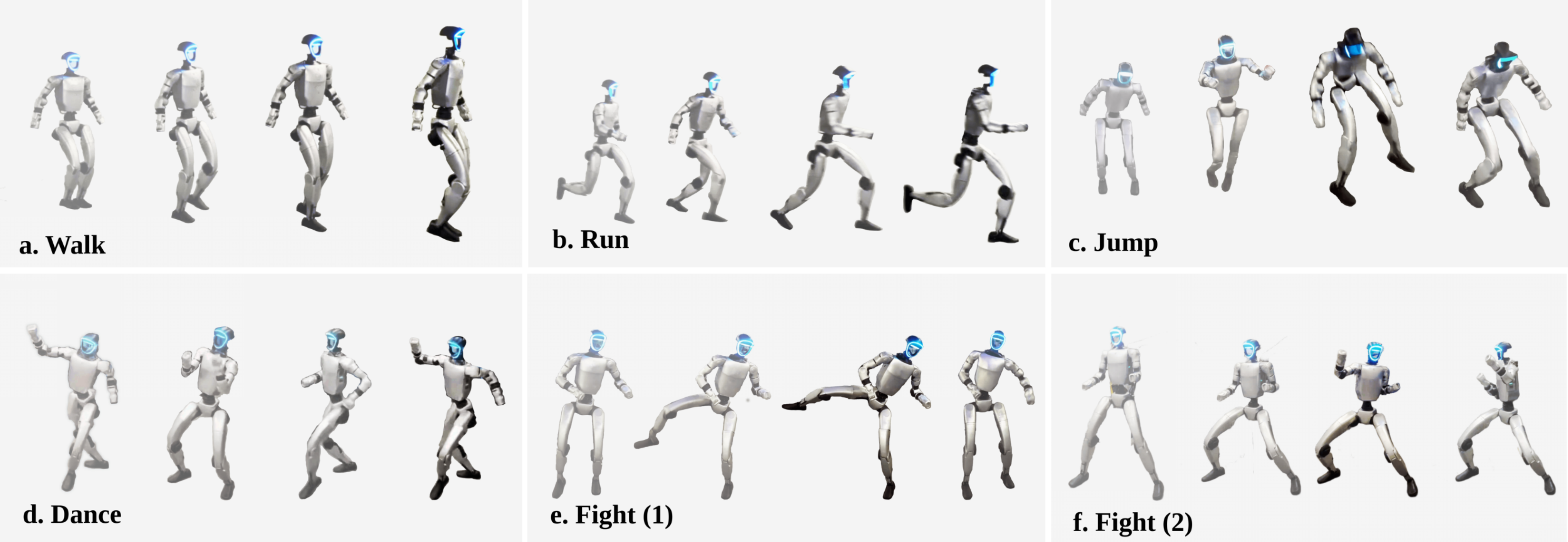}
\caption{Representative whole-body motion skills (a subset of the six skill categories) learned sequentially in a continual-learning setting, where the policy must acquire new behaviors while retaining previously learned locomotion and athletic motion skills}
\label{fig:teaser}
\end{figure}

\section{Introduction}
\label{sec:intro}
Whole-body control underpins humanoid behavior: robust locomotion requires coordinating the legs, torso, and arms to manage balance and momentum while supporting task-relevant motion. Driven by learning-based motion tracking with sim-to-real pipelines that exploit large-scale human motion, a single policy can track or generalize across a broad repertoire of agile whole-body skills~\cite{cheng2024expressive,he2025asap,radosavovic2024real,luo2024universal,han2025kungfubot2}. Yet such policies are trained offline and deployed as essentially static controllers, making it hard to \textit{incrementally} acquire new motions. The obvious workaround---fine-tuning on the new motion set---triggers catastrophic forgetting, often causing severe regressions on previously mastered behaviors~\cite{wang2024comprehensive,zhou2024continual}. This contrasts sharply with human learning, where new skills are added while old ones are retained. Continual learning (CL) explicitly targets this stability--plasticity dilemma and is well studied in computer vision and natural language processing~\cite{shi2025continual}, but in robotics it remains under-explored and is demonstrated mostly in manipulation and navigation rather than legged whole-body control~\cite{mendonca2025continuously,wan2024lotus,liu2024semi}. To our knowledge, continual learning for humanoid whole-body motion has not been systematically studied. Such continual acquisition can also extend generalist controllers~\cite{luo2025sonic}, offering a practical route toward an ever-growing generalist.

We therefore target this gap: enabling one policy to \textit{sequentially} acquire whole-body skills (Fig.~\ref{fig:teaser}) with strong transfer at low cost. Building on Progressive Neural Networks (PNN)~\cite{rusu2016progressive}, we instantiate a frozen sub-network per task, which isolates prior competencies and makes zero forgetting a structural guarantee rather than something we must optimize for. The challenge, and our focus, is to make this expansion reuse knowledge across skills while staying parameter- and computation-efficient. We grow each new sub-network with parameter-efficient LoRA adapters~\cite{hu2022lora} instead of training it from scratch, and we guide expansion with a two-level motion-similarity measure that selects the most related prior skill to inherit from and sets each adapter's capacity from the expected transfer. Across a continual suite of six skill categories, this yields the best forward transfer among strong baselines, while cutting trainable parameters and training time substantially, with zero forgetting by design.

\bmhead{Contributions}
\begin{itemize}
\setlength{\itemsep}{1pt}\setlength{\topsep}{2pt}\setlength{\parsep}{0pt}
\item We formulate and benchmark \emph{continual humanoid whole-body motion learning}, providing the first pipeline for continually acquiring humanoid whole-body skills, together with adapted multi-motion and standard-CL baselines under a unified protocol.
\item We propose \emph{Similarity-guided LoRA-PNN}, in which a motion-similarity measure drives inheritance from the most related prior skill and adaptive expansion, making continual learning both parameter- and computation-efficient.
\item Our method delivers the strongest forward transfer and large efficiency gains (up to $94.5\%$ fewer trainable parameters and $40.8\%$ less training time) with zero catastrophic forgetting, validated through sim-to-sim transfer and real-robot deployment on a Unitree~G1.
\end{itemize}

\section{Related Work}
\label{sec:related}
\bmhead{Humanoid whole-body locomotion}
Learning-based control has rapidly expanded the repertoire of humanoid whole-body skills, from real-world RL locomotion driven by proprioceptive history~\cite{radosavovic2024real} and expressive upper-/lower-body coordination~\cite{cheng2024expressive} to unified controllers that expose fine-grained command spaces~\cite{xue2025unified}. A particularly active direction is whole-body \emph{motion tracking}, where a single policy imitates large libraries of reference motions~\cite{he2025asap,han2025kungfubot2} and, more recently, scales to generalist controllers that track thousands of clips~\cite{luo2025sonic}. Despite this progress, such controllers are typically trained offline and then deployed as fixed models: incorporating a new skill requires retraining or fine-tuning, which degrades previously mastered behaviors---in stark contrast to how humans accumulate skills over time.

\bmhead{Continual learning}
Continual (or lifelong) learning studies how an agent can incrementally acquire tasks from a non-stationary stream while balancing the stability--plasticity trade-off and mitigating catastrophic forgetting~\cite{wang2024comprehensive}. Classical strategies fall into three families: regularizing important weights (\eg, EWC~\cite{kirkpatrick2017overcoming}), parameter isolation via pruning or masking (\eg, PackNet~\cite{mallya2018packnet}), and dynamic architecture growth such as Progressive Neural Networks~\cite{rusu2016progressive}; more recent work adapts large foundation models with lightweight low-rank modules to limit interference~\cite{luos2026keeplora}. In robotics, continual learning has been explored mainly for manipulation and mobile platforms---\eg, continually improving mobile manipulation~\cite{mendonca2025continuously}, growing skill libraries through unsupervised discovery~\cite{wan2024lotus}, online neurogenesis for lifelong legged control~\cite{srisuchinnawong2025growable}, self-evolving VLA skill spaces with expert routing~\cite{wu2025continually}, and language-augmented skill composition~\cite{meng2025preserving}. To the best of our knowledge, continual learning has not been investigated for humanoid whole-body motion. We address this gap with \emph{Similarity-guided LoRA-PNN}, a progressive policy that continually acquires humanoid whole-body skills with zero forgetting and strong, low-cost transfer.

\section{Method}
\label{sec:method}

\subsection{Problem Setup}
\label{subsec:method_setup}
We study continual learning for humanoid whole-body motion, where a single controller must incrementally acquire a growing set of whole-body skills (\eg, walking, jumping, and dancing) from a \textit{sequential} curriculum in which skills are introduced one after another. Let $\pi_{\theta}$ map humanoid observations (proprioception, root state, contact cues, and optional task commands) to actions, and let $\{\mathcal{D}_1,\ldots,\mathcal{D}_K\}$ be the ordered task stream, where $\mathcal{D}_k$ holds the demonstrations for task $k$. A continual-learning algorithm $\mathcal{A}_{\mathrm{CL}}$ updates the network online across skills via
\begin{equation}
\theta_k \;=\; \mathcal{A}_{\mathrm{CL}}(\theta_{k-1},\mathcal{D}_k)\enspace,
\label{eq:setup}
\end{equation}
seeking \textit{plasticity} on the newly introduced behavior while preserving \textit{stability} on previously learned ones. Our approach removes the stability bottleneck altogether, so that mastering a new skill never degrades those already learned. This shifts the central question from \emph{whether the controller forgets} to \emph{how quickly it transfers to new skills and how efficiently it scales} as its repertoire grows.

\subsection{Continual Learning Policy}
\label{subsec:cl_policy}
Each task $\mathcal{T}_k$, $k\in\{1,\dots,K\}$, requires a humanoid to learn skill $k$ from demonstrations $\mathcal{D}_k$. Following prior whole-body tracking work~\cite{cheng2024expressive,he2025omnih2o,xie2025kungfubot,han2025kungfubot2}, we cast each task as \textit{goal-conditioned reinforcement learning} in joint space: at step $t$, an actor $\pi_\theta$ receives the humanoid state $s_t$ and a reference state $g_t$ and outputs a continuous action $a_t\in\mathbb{R}^{23}$ (for the Unitree G1), which a PD controller converts to torques. With transitions $p(s_{t+1}\mid s_t,a_t)$ and a dense tracking-and-regularization reward $r(s_t,g_t,a_t)$, we maximize the expected discounted return:
\begin{equation}
    J(\theta)=\mathbb{E}\Big[\sum_{t=0}^{T-1}\gamma^t\, r(s_t,g_t,a_t)\Big], \quad a_t \sim \pi_\theta(\cdot\mid s_t,g_t)\enspace,
\end{equation}
using PPO~\cite{schulman2017proximal} with a goal-conditioned critic $V_\phi(s_t,g_t)$ that provides low-variance learning signals.

\begin{figure}[!htb]
    \centering
    \includegraphics[width=\textwidth]{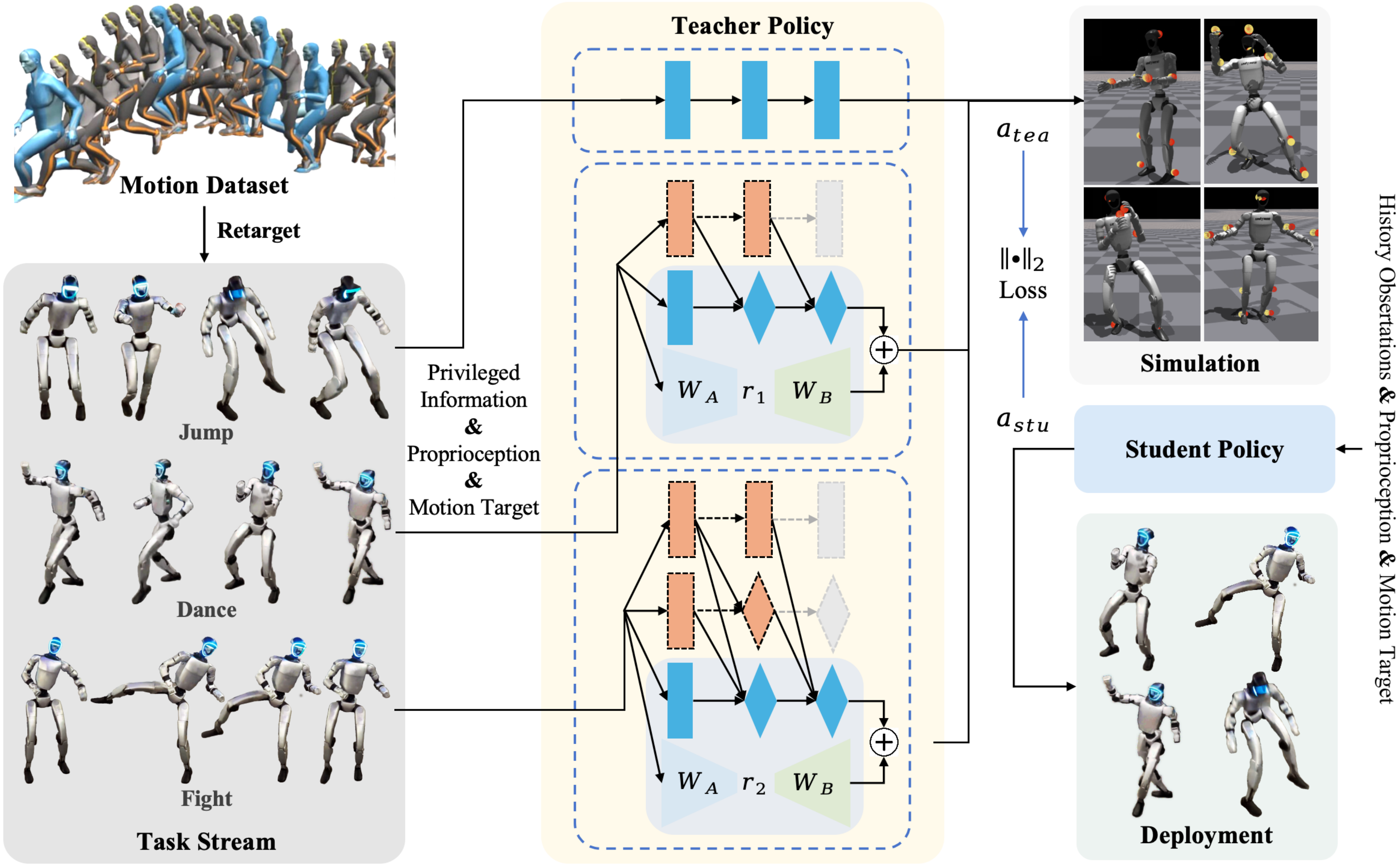}
    \caption{Overview of the continual whole-body locomotion learning pipeline. A collection of retargeted motions~\cite{harvey2020robust} is organized into a sequential stream of tasks (\eg, jumping, dancing, and fighting) and learned using a teacher--student framework. The teacher policy, with access to privileged information, is expanded for each new task by adding a trainable column (solid blue) via similarity-guided LoRA, while previously learned columns (dashed orange) are kept frozen to preserve earlier skills. The teacher's actions supervise the student policy through a distillation loss, and the student is deployed in the real world}
    \label{fig:overview}
\end{figure}

Once a policy is trained for skill $k$, it is typically frozen, and naively fine-tuning it on a new skill induces catastrophic forgetting; joint multi-skill training~\cite{luo2023perpetual,luo2024universal,he2025omnih2o,han2025kungfubot2,shi2025adversarial} likewise cannot absorb additional skills afterwards without degrading earlier ones. To overcome this, we build a continual policy on Progressive Neural Networks (PNN)~\cite{rusu2016progressive}, which promote transfer while preventing interference. As illustrated in Fig.~\ref{fig:overview}, we instantiate a separate actor ``column'' per task and \textit{freeze} all earlier columns, eliminating inter-task interference in optimization. Each column is a multilayer perceptron (MLP) with lateral connections; after training the first column to convergence, learning task $k$ adds a fresh column whose hidden layer receives within-column input plus \textit{lateral inputs} from earlier columns~\cite{rusu2016progressive}:
\begin{equation}
h_i^{(k)} \;=\; f\!\left(W_i^{(k)}\, h_{i-1}^{(k)} \;+\; \sum_{j<k} U_i^{(k:j)}\, h_{i-1}^{(j)}\right)\enspace,
\end{equation}
where $W_i^{(k)}$ are the feedforward weights of column $k$, $U_i^{(k:j)}$ are lateral connections from column $j\!<\!k$, and $f(\cdot)$ is an element-wise nonlinearity. Transfer thus occurs \textit{at every layer}, letting the new column reuse, modify, or ignore prior representations, while the directed connectivity and parameter freezing guarantee that columns $1,\dots,k-1$ remain unchanged, which rules out forgetting by construction. All columns share a single critic, which is discarded after training.

\subsection{Parameter-Efficient Policy Learning}
\label{subsec:lora}
PNN avoids forgetting by freezing prior columns, but its parameter count grows with the number of tasks, adding training overhead as tasks accumulate. Instead of randomly initializing each new column and training it from scratch~\cite{rusu2016progressive}, we propose a \emph{motion-similarity--guided LoRA adaptation} strategy for every column after the first. Given demonstration sets $\mathcal{D}_l$ and $\mathcal{D}_k$ ($1\le l,k\le K$), we precompute a set-level motion similarity $S(l,k)=\exp\!\big(-D_{\text{set}}(\mathcal{D}_l,\mathcal{D}_k)/\tau_s\big)\in(0,1]$ from the motion-set distance defined in Sec.~\ref{subsec:sim} (larger $S$ means more similar), and proceed in two stages: parent-column inheritance and similarity-guided rank selection.

\bmhead{Parent column inheritance}
When task $k$ arrives, we select its parent column from the precomputed similarities,
\begin{equation}
p(k) = \arg\max_{l<k} S(l,k)\enspace,
\label{eq:max_sim}
\end{equation}
\ie, the previously learned column whose motions are most kinematically similar to task $k$. We initialize column $k$ by copying the parameters of $p(k)$ (instead of random initialization~\cite{rusu2016progressive}) and then optimize \emph{only} the parameters of task $k$. This warm start supplies a stronger inductive bias when the incoming motions share structure with earlier ones, improving learning efficiency and stability.

\begin{table}[!htb]
    \centering
    \caption{\textbf{LoRA rank allocation strategies.} The allowed ranks are $r_{\min},r_{\mathrm{mid}},r_{\max}\in\{4,8,16\}$; $\mathrm{normalize}$ min--max rescales similarity to $[0,1]$ over the candidate parents $\{l<k\}$; $\gamma_1,\gamma_2,\tau_s,\alpha$ are fixed (see text)}
    \label{tab:lora_rule}
    \setlength{\tabcolsep}{12pt}
    \begin{tabular}{ll}
    \toprule
    \textbf{Strategy} & \textbf{Rank Rule} \\
    \midrule
    Fixed &
    $
    r_k = r_1 \qquad \forall\, 1 \leq k \leq K\enspace.
    $ \\
    \midrule
    Binned &
    $
    \begin{aligned}
    & \bar S_k \coloneqq \mathrm{normalize}\big(S(p(k), k)\big) \in [0,1] \enspace, \\
    & r_k=
    \begin{cases}
    r_{\min}, & \bar S_k \ge \gamma_1 \enspace, \\
    r_{\text{mid}}, & \gamma_2 \le \bar S_k < \gamma_1 \enspace,\\
    r_{\max}, & \bar S_k < \gamma_2 \enspace.
    \end{cases}
    \end{aligned}
    $
    \\
    \midrule
    Linear &
    $
    \begin{aligned}
    & \bar S_k \coloneqq \mathrm{normalize}\big(S(p(k), k)\big) \in [0,1] \enspace, \\
    & r_k = \left\lfloor
    \mathrm{clip}\!\left(
    r_{\max}-\alpha\,\bar S_k,\, r_{\min},\, r_{\max}
    \right)
    \right\rceil\enspace.
    \end{aligned}
    $
    \\
    \bottomrule
    \end{tabular}
\end{table}

\bmhead{Similarity-guided LoRA rank selection}
To make adaptation parameter- and time-efficient, we parameterize each linear layer of every non-first column with LoRA~\cite{hu2022lora}: for a base weight $W\in\mathbb{R}^{d_{\text{out}}\times d_{\text{in}}}$ cloned from the parent, we keep $W$ frozen and learn a low-rank update
\begin{equation}
W^\prime = W + \Delta W\enspace, \quad \Delta W = W_BW_A\enspace,
\label{eq:lora}
\end{equation}
with $W_A\in\mathbb{R}^{r\times d_{\text{in}}}$, $W_B\in\mathbb{R}^{d_{\text{out}}\times r}$, and $r \ll \min(d_{\text{in}}, d_{\text{out}})$. We further modulate adaptation capacity by similarity: when task $k$ closely matches its parent $p(k)$, a lower-rank update suffices, so we set $r_k=f(S(p(k),k))$ with $\partial f/\partial S<0$. Strong-transfer tasks thus receive smaller ranks for faster training, while dissimilar tasks receive larger ranks for more capacity. We discretize $\{r_k\}_{k=2}^K$ into a small allowed set and use the three rules in Table~\ref{tab:lora_rule}. All LoRA hyperparameters are fixed across tasks: similarity temperature $\tau_s{=}1.0$ ($1.5$ for the binned rule), $\gamma_1{=}0.7$, $\gamma_2{=}0.4$, linear slope $\alpha{=}12.0$, ranks $r\in\{4,8,16\}$, LoRA scaling $\alpha_{\text{LoRA}}{=}1.0$, and no dropout.

\subsection{Motion Similarity}
\label{subsec:sim}
Parent selection and rank allocation both rely on a distance between two \emph{sets} of motions, $\mathcal{D}_l=\{A_n\}_{n=1}^N$ and $\mathcal{D}_k=\{B_m\}_{m=1}^M$, where each motion is a variable-length sequence (typically 3--5 minutes after retargeting) that may contain transitions (pauses, start/stop, activity changes). The distance must be robust to (i) temporal misalignment and (ii) unequal sequence lengths, and it must not be driven by how many motions each set happens to contain. We therefore use a \textit{two-level comparison}: motions are first decomposed into short windows and compared as \emph{distributions} of local behaviors, and sets are then compared as distributions of motions. Similarity is computed hierarchically with dynamic time warping (DTW)~\cite{senin2008dynamic} inside windows and optimal transport (OT)~\cite{su2017order} across windows and across motions.

\bmhead{Per-frame features}
All motions are resampled to a common rate $f_0$ ($\Delta t=1/f_0$) and converted into translation-invariant, heading-normalized features. Let $p_t=(x_t,y_t)$ be the planar root position and $\psi_t$ the root yaw at frame $t$. We use the local-frame planar velocity, the yaw rate, and the per-joint incremental rotation magnitude,
\begin{equation}
\begin{aligned}
v_t &= R(-\psi_t)\,\frac{p_{t+1}-p_t}{\Delta t}\enspace, \qquad
\omega_t = \frac{\mathrm{wrap}(\psi_{t+1}-\psi_t)}{\Delta t}\enspace, \\
\alpha_{t,j} &= d_{SO(3)}\!\big(R_{t,j}^\top R_{t+1,j},\, I\big)\enspace, \qquad j\in\mathcal{J}\enspace,
\end{aligned}
\label{eq:features}
\end{equation}
where $R(\cdot)$ is the 2D rotation matrix, $\mathrm{wrap}$ maps angles to $(-\pi,\pi]$, $R_{t,j}\in SO(3)$ is the rotation of joint $j$, and $d_{SO(3)}$ is the geodesic (rotation-angle) distance, so that $\alpha_{t,j}\in[0,\pi]$ measures how much joint $j$ moves between consecutive frames. Together with an optional contact signal $c_t\in\mathbb{R}^K$, treated as real-valued because contact labels may be soft, the per-frame feature is $f_t=[v_t,\,\omega_t,\,\{\alpha_{t,j}\}_{j\in\mathcal{J}},\,c_t]$. These features are invariant to global translation and largely invariant to global heading, which matters for long sequences whose absolute positions drift across transitions.

\bmhead{Windowing}
Each motion is decomposed into overlapping windows of duration $L$ seconds and stride $S$ seconds ($L_f=Lf_0$ and $S_f=Sf_0$ frames). The $k$-th window of motion $A$ and its normalized center time are
\begin{equation}
W^A_k=\{f_t\}_{t=t_k}^{t_k+L_f-1}\enspace,\qquad
\tau^A_k=\frac{t_k+L_f/2}{T_A-1}\in[0,1]\enspace,
\label{eq:window}
\end{equation}
with $t_k=kS_f$ and $T_A$ the number of frames of $A$. This converts a long, transition-laden sequence into a set of short local behaviors (steps, turns, pauses) that can be matched individually.

\bmhead{Window cost via DTW}
For a window pair $(W^A_k,W^B_\ell)$ we define a weighted per-frame cost between local frames $t$ and $s$,
\begin{equation}
\delta(t,s)= w_v\|v^A_t-v^B_s\|_2 + w_\omega|\omega^A_t-\omega^B_s|
+ \frac{w_J}{|\mathcal{J}|}\sum_{j\in\mathcal{J}}|\alpha^A_{t,j}-\alpha^B_{s,j}|
+ \frac{w_c}{K}\|c^A_t-c^B_s\|_1\enspace,
\label{eq:frame_cost}
\end{equation}
and align the two windows by DTW under a Sakoe--Chiba band $|t-s|\le b$~\cite{sakoe2003dynamic} that prevents pathological warps. With $\pi^\star$ the optimal warping path, the window-to-window cost is the path-length-normalized DTW cost
\begin{equation}
c_{k\ell}=\frac{1}{|\pi^\star|}\sum_{(t,s)\in\pi^\star}\delta(t,s)\enspace,
\label{eq:window_cost}
\end{equation}
which absorbs the timing differences that arise even when different actors perform the same motion.

\bmhead{Motion-to-motion distance}
To compare two entire motions we match their windows as distributions rather than forcing a single global alignment across minutes. Each window cost is augmented with a soft temporal-order penalty, $\tilde c_{k\ell}=c_{k\ell}+\beta\,|\tau^A_k-\tau^B_\ell|$ with $\beta\ge 0$, and the windows are matched by an OT problem with uniform masses:
\begin{equation}
D_{\text{motion}}(A,B) = \min_{P\ge 0}\sum_{k,\ell} P_{k\ell}\,\tilde c_{k\ell}
\quad \text{s.t.} \quad P\mathbf{1}=a\enspace,\; P^\top\mathbf{1}=b\enspace,
\label{eq:dmotion}
\end{equation}
where $P$ is the transport plan and $a_k=1/n_A$, $b_\ell=1/n_B$ for window counts $n_A,n_B$. The entropic form of this problem is solved with Sinkhorn iterations~\cite{cuturi2013sinkhorn}. Because low-motion windows can be matched to other low-motion windows at low cost, the distance is robust to pauses and transitions.

\bmhead{Set-to-set distance}
A second, category-level OT over the motion--motion cost matrix $C_{ij}=D_{\text{motion}}(A_i,B_j)$ gives the set-to-set distance
\begin{equation}
D_{\text{set}}(\mathcal{D}_l,\mathcal{D}_k) = \min_{Q\ge 0}\sum_{i,j}Q_{ij}C_{ij}\quad \text{s.t.} \quad Q\mathbf{1}=u\enspace,\; Q^\top\mathbf{1}=v\enspace,
\label{eq:dset}
\end{equation}
with uniform masses $u_i=1/N$ and $v_j=1/M$. Because both stages use uniform weights, the OT objectives compare empirical distributions rather than raw counts: the resulting scalar reflects differences in \emph{motion patterns and their coverage}, not the set sizes $N$ and $M$. For robustness across segmentation scales, distances are averaged over four window/stride configurations. The motion data format, the window/stride configurations, and the remaining implementation constants are listed in Appendix~\ref{app:sim}.

\begin{figure}[t]
  \centering
  \includegraphics[width=.55\textwidth]{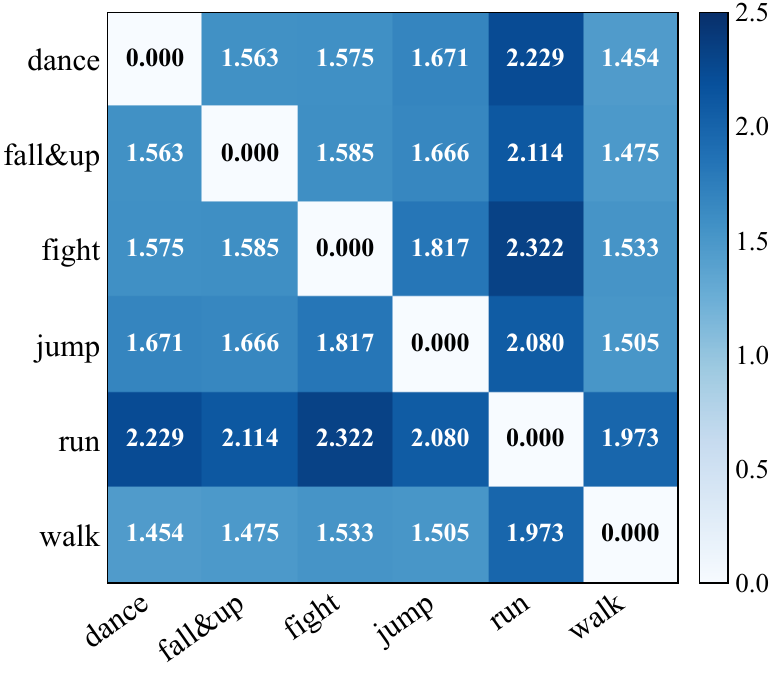}
  \caption{\textbf{Motion-set distance matrix} over the six categories (lighter\,=\,more similar)}
  \label{fig:sim_heatmap}
\end{figure}

\bmhead{Resulting distance matrix}
Fig.~\ref{fig:sim_heatmap} shows the pairwise distances among the six categories. Two trends stand out. \texttt{run} is the most distinctive skill, with the largest distance to every other task (\eg, \texttt{run}--\texttt{fight} $=2.32$ and \texttt{run}--\texttt{dance} $=2.23$): high-speed locomotion produces motion patterns that transfer least to the remaining skills. \texttt{walk}, in contrast, is central, with consistently small distances to several tasks (\texttt{walk}--\texttt{dance} $=1.45$, \texttt{walk}--\texttt{fall\_and\_up} $=1.48$), reflecting shared periodic stepping and lower-body coordination. Among the non-running behaviors the distances lie in a narrow band ($1.56$--$1.82$). The matrix thus separates locomotion-dominant tasks from more diverse whole-body skills and gives a quantitative basis for parent selection (Eq.~\ref{eq:max_sim}) and rank allocation (Table~\ref{tab:lora_rule}). It is computed once offline before training, at negligible cost relative to RL optimization, and is excluded from the per-task time budget.

\subsection{Policy Training Objective}
\label{sec:objective}
We train each column as a distinct actor with an on-policy objective that imitates the reference motion: at stage $k$ only column $k$ is optimized while all earlier columns are frozen, no past-task replay is used, and all actors share a single critic. The per-step reward is a weighted sum of tracking and regularization terms:
\begin{equation}
R \;=\; \sum_{i\in\mathcal{I}} w_i \, r_i \;+\; \sum_{j\in\mathcal{J}} \lambda_j \, p_j\enspace,
\label{eq:reward_sum}
\end{equation}
where $\mathcal{I}$ indexes tracking terms and $\mathcal{J}$ penalty terms. Following the adaptive scheme of~\cite{xie2025kungfubot}, each tracking term is an exponential of a normalized error, $r_i=\exp(-\lVert e_i\rVert/\sigma_i)$, covering root-velocity, anchor (key-body) position/orientation, and key-body linear/angular-velocity tracking, with the same terms and weights shared across all tasks. Regularizers~\cite{cheng2024expressive} penalize action rate, joint velocity/acceleration, foot slip, and limit/collision violations, and episodes terminate when the reference motion ends or safety constraints are violated.

\bmhead{Teacher--student distillation}
\label{paragraph:teacher_student}
To obtain a deployable policy, we distill a privileged teacher into a student that uses only deployable observations. The teacher is trained in Isaac Gym~\cite{makoviychuk2021isaac} with privileged inputs (base velocity, key-body poses, contact states, and domain-randomization parameters) unavailable in MuJoCo~\cite{todorov2012mujoco} and on hardware. The student is trained with DAgger~\cite{ross2011reduction}, predicting actions from deployable observations augmented with a history encoder while minimizing $\|a_\text{tea}-a_\text{stu}\|_2$. Both policies use a shared encoder over the deployable proprioceptive history; this encoder is trained jointly with the teacher and then kept frozen during student distillation, so distillation only updates the student's action prediction network. Table~\ref{tab:obs_teacher_student} lists the observation design of both policies; every input the student consumes is available on the physical Unitree~G1.

\begin{table}[!htb]
\centering
\caption{\textbf{Observation design for teacher--student distillation.} The teacher is trained with privileged observations; the student uses only deployable observations plus a history encoding}
\label{tab:obs_teacher_student}
\setlength{\tabcolsep}{3pt}
\begin{tabular}{>{\raggedright\arraybackslash}p{0.30\textwidth} c c c >{\raggedright\arraybackslash}p{0.31\textwidth}}
\toprule
\textbf{Observation / signal} & \textbf{Teacher} & \textbf{Student} & \textbf{Deployable} & \textbf{Notes} \\
\midrule
\multicolumn{5}{l}{\textit{Proprioception (common, deployable)}} \\
\texttt{dof\us pos} & \checkmark & \checkmark & \checkmark & Joint positions \\
\texttt{dof\us vel} & \checkmark & \checkmark & \checkmark & Joint velocities \\
\texttt{actions} & \checkmark & \checkmark & \checkmark & Previous action \\
\texttt{next\us step\us ref\us motion} & \checkmark & \checkmark & \checkmark & Reference-motion conditioning (available in deployment when tracking a given motion) \\
\texttt{history} & & \checkmark & \checkmark & Past $K{=}10$ frames of $[\texttt{dof\us pos},\texttt{dof\us vel},\texttt{actions}]$ \\
\midrule
\multicolumn{5}{l}{\textit{Privileged, simulation-only (teacher only)}} \\
\texttt{base\us lin\us vel} & \checkmark & & \texttimes & Base linear velocity \\
\texttt{base\us ang\us vel} & \checkmark & & \texttimes & IMU-dependent; removed for the deployable student \\
\texttt{roll\us pitch} & \checkmark & & \texttimes & IMU-dependent base-orientation cues \\
\texttt{local\us key\us body\us pos} & \checkmark & & \texttimes & Key-body local positions \\
\texttt{local\us key\us body\us rot} & \checkmark & & \texttimes & Key-body local rotations \\
\texttt{anchor\us ref\us pos} & \checkmark & & \texttimes & Anchor reference position \\
\texttt{anchor\us ref\us rot} & \checkmark & & \texttimes & Anchor reference rotation \\
\texttt{contact\us mask} & \checkmark & & \texttimes & Contact state (privileged input / critic signal) \\
\texttt{root\us height} & \checkmark & & \texttimes & Root height \\
\texttt{dr\us *} & \checkmark & & \texttimes & Domain-randomization parameters (mass, friction, gains, \etc) \\
\midrule
\multicolumn{5}{l}{\textit{Future-motion targets (conditioning)}} \\
\texttt{future\us motion\us root\us height} & \checkmark & \checkmark & \checkmark & Future reference target \\
\texttt{future\us motion\us roll\us pitch} & \checkmark & \checkmark & \checkmark & Future reference target \\
\texttt{future\us motion\us base\us lin\us vel} & \checkmark & \checkmark & \checkmark & Derived from the reference trajectory \\
\texttt{future\us motion\us base\us yaw\us vel} & \checkmark & \checkmark & \checkmark & Future reference target \\
\texttt{future\us motion\us dof\us pos} & \checkmark & \checkmark & \checkmark & Future joint targets from the reference motion \\
\texttt{future\us motion\us local\us ref\us key\us body\us pos} & \checkmark & & \checkmark & Optional; used by the teacher, omitted by the student \\
\bottomrule
\end{tabular}
\end{table}

\section{Experiments}
\label{sec:exp}

\subsection{Experimental Setup}
\label{subsec:exp_setup}

\bmhead{Task suite and continual protocol}
We study continual humanoid whole-body motion learning under a task-incremental protocol with a provided task identifier. Let $\mathcal{K}=\{1,\dots,K\}$ be the task set, trained in order $1\!\rightarrow\!2\!\rightarrow\!\cdots\!\rightarrow\!K$. After learning task $k$, we evaluate the policy $\pi_k$ on all tasks $i\le k$, forming a stage-by-task matrix $A\in\mathbb{R}^{K\times K}$ with $A_{k,i}=\mathrm{Perf}(\pi_k,i)$. Our suite contains $K=6$ tasks curated from \textbf{LAFAN1}~\cite{harvey2020robust} and the \textbf{Kungfu} motion datasets~\cite{xie2025kungfubot,han2025kungfubot2} to span a range of difficulty, from easiest to hardest: \texttt{walk}, \texttt{run}, \texttt{jump}, \texttt{dance}, \texttt{fight}, and \texttt{fall\_and\_up}. All methods share the same per-task optimization budget (50K iterations, 1024 parallel environments) and evaluation budget (100 episodes per task). Key implementation settings are summarized in Table~\ref{tab:hparams}; the complete PPO, network, domain-randomization, and reward settings are listed in Appendix~\ref{app:hparams}.

\begin{table}[t]
\centering
\setlength{\tabcolsep}{4pt}
\caption{\textbf{Key implementation hyperparameters.} Full reward/termination weights and domain-randomization ranges follow standard whole-body tracking practice~\cite{cheng2024expressive,xie2025kungfubot,han2025kungfubot2}}
\label{tab:hparams}
\begin{tabular}{@{}l>{\raggedright\arraybackslash}p{0.70\textwidth}@{}}
\toprule
Networks & actor MLP $[512,512,256]$, critic $[768,512,256]$ (LayerNorm), Conv1D encoder $128/64$, SiLU \\
PPO & lr${=}1\mathrm{e}{-4}$, clip${=}0.2$, $\gamma{=}0.99$, GAE${=}0.95$, entropy${=}0.01$, $24$ steps, $4$ mini-batches, $5$ epochs \\
Budget & $50$K iterations, $1024$ envs, $100$ evaluation episodes per task \\
LoRA & $(r_{\min},r_{\mathrm{mid}},r_{\max}){=}(4,8,16)$, $\alpha_{\text{LoRA}}{=}1.0$, dropout${=}0$ \\
Rank rules & $\gamma_1{=}0.7$, $\gamma_2{=}0.4$, $\alpha{=}12.0$, $\tau_s{=}1.0$ ($1.5$ binned) \\
Sim.\ windows & $(L,S)\in\{(3,2),(5,3),(10,7),(15,10)\}$\,s \\
\bottomrule
\end{tabular}
\end{table}

\bmhead{Metrics}
From the score matrix $A$ we report six continual-learning metrics: average accuracy (\emph{AA}) and average incremental accuracy (\emph{AIA}) measure overall and trajectory-averaged performance over learned tasks; the forgetting measure (\emph{FM}) and backward transfer (\emph{BWT}) measure degradation on earlier tasks; intransigence (\emph{IM}) measures the gap to a joint-training reference; and forward transfer (\emph{FWT}) measures the benefit on a newly introduced task relative to a single-task-from-scratch reference. Table~\ref{tab:cl_metrics_def} gives the formal definitions in terms of $A$. We instantiate $\mathrm{Perf}$ with two scores: success rate and tracking accuracy, for which we use a $-$MPJPE (negative mean per-joint position error) score~\cite{han2025kungfubot2,xie2025kungfubot,he2025asap,luo2024universal}; an episode is a success if the policy tracks the reference motion to completion without early termination (no fall and within the training root-deviation and body-height safety bounds). For FWT, $A^{\mathrm{ST}}_i$ is the score of a policy trained from scratch on task $i$, and $A_{i-1,i}$ uses the model available just before task $i$ is learned; for IM, $A^{\mathrm{JT}}_i$ is the score of a policy jointly trained on all tasks. We additionally report the imitation errors ($E_{\text{mpjpe}}$, $E_{\text{gmpjpe}}$, $E_{\text{mpbpe}}$, $E_{\text{mpjve}}$, $E_{\text{mpjae}}$, $E_{\text{pbve}}$, $E_{\text{pbae}}$)~\cite{han2025kungfubot2,xie2025kungfubot,he2025asap,luo2024universal} as tracking-quality sanity checks; Table~\ref{tab:mimic_metric_defs} defines them.

\begin{table}[!htb]
\centering
\caption{\textbf{Definitions of the continual-learning metrics.} $A_{t,i}$ denotes the score of the policy after learning tasks $1{:}t$, evaluated on task $i$; $A^{\mathrm{JT}}_i$ and $A^{\mathrm{ST}}_i$ are the scores of a joint-trained reference and of a single-task reference trained only on task $i$}
\label{tab:cl_metrics_def}
\setlength{\tabcolsep}{4pt}
\begin{tabular}{l >{\raggedright\arraybackslash}p{0.42\textwidth} >{\raggedright\arraybackslash}p{0.40\textwidth}}
\toprule
\textbf{Metric} & \textbf{Description} & \textbf{Definition} \\
\midrule
AA & Average accuracy over the tasks learned so far (overall continual performance after stage $t$) &
$\displaystyle \mathrm{AA}_t = \frac{1}{t}\sum_{i=1}^{t} A_{t,i}$ \\[6pt]
AIA & Average incremental accuracy: the historical average of AA up to stage $t$ (quality of the whole learning trajectory) &
$\displaystyle \mathrm{AIA}_t = \frac{1}{t}\sum_{k=1}^{t} \mathrm{AA}_k$ \\[6pt]
FM & Forgetting measure: how much performance on previous tasks degrades after learning new tasks, relative to the historical best &
$\displaystyle \mathrm{FM}_t = \frac{1}{t-1}\sum_{i=1}^{t-1}\Big(\max_{k\in\{i,\dots,t-1\}} A_{k,i} - A_{t,i}\Big)$ \\[6pt]
BWT & Backward transfer: net change on past tasks after learning subsequent tasks; negative values indicate forgetting &
$\displaystyle \mathrm{BWT}_t = \frac{1}{t-1}\sum_{i=1}^{t-1}\big(A_{t,i} - A_{i,i}\big)$ \\[6pt]
IM & Intransigence: gap to the joint-training reference, \ie, how far the continual learner is from the multi-task upper bound &
$\displaystyle \mathrm{IM}_t = \frac{1}{t}\sum_{i=1}^{t}\big(A^{\mathrm{JT}}_i - A_{t,i}\big)$ \\[6pt]
FWT & Forward transfer: benefit (or harm) on a new task before learning it, relative to a single-task reference trained from scratch &
$\displaystyle \mathrm{FWT}_t = \frac{1}{t-1}\sum_{i=2}^{t}\big(A_{i-1,i} - A^{\mathrm{ST}}_i\big)$ \\
\bottomrule
\end{tabular}
\end{table}

\begin{table}[!htb]
\centering
\caption{\textbf{Definitions of the imitation metrics.} $t\in\{1,\dots,T\}$ indexes frames, $j\in\{1,\dots,J\}$ joints, $r$ the root (pelvis), and $b\in\mathcal{B}$ body keypoints (\eg, torso, feet, hands). Predicted and reference positions are $\mathbf{p}_{t,j},\hat{\mathbf{p}}_{t,j}\in\mathbb{R}^3$ (metres, converted to mm by the factor $10^3$)}
\label{tab:mimic_metric_defs}
\setlength{\tabcolsep}{3pt}
\begin{tabular}{l c >{\raggedright\arraybackslash}p{0.29\textwidth} l}
\toprule
\textbf{Metric} & \textbf{Unit} & \textbf{Meaning} & \textbf{Formula} \\
\midrule
Succ.\ $\uparrow$ & \% & Success rate over evaluation episodes (episode finishes without failure or early termination) &
$\frac{1}{N}\sum_{n=1}^N \mathbb{I}[\text{episode $n$ succeeds}]$ \\[4pt]
$E_{\text{mpjpe}}\downarrow$ & mm & Mean per-joint position error, root-relative (global translation removed by aligning the root) &
\makecell[l]{$\frac{10^3}{TJ}\sum_{t,j}\big\lVert (\mathbf{p}_{t,j}-\mathbf{p}_{t,r})$\\[2pt] $\quad -(\hat{\mathbf{p}}_{t,j}-\hat{\mathbf{p}}_{t,r})\big\rVert_2$} \\[4pt]
$E_{\text{gmpjpe}}\downarrow$ & mm & Global MPJPE: world-frame joint position error, no root alignment &
$\frac{10^3}{TJ}\sum_{t,j}\big\lVert \mathbf{p}_{t,j}-\hat{\mathbf{p}}_{t,j}\big\rVert_2$ \\[4pt]
$E_{\text{mpbpe}}\downarrow$ & mm & Mean per-body keypoint position error over a predefined set $\mathcal{B}$ (key bodies / end-effectors) &
$\frac{10^3}{T|\mathcal{B}|}\sum_{t}\sum_{b\in\mathcal{B}}\big\lVert \mathbf{x}_{t,b}-\hat{\mathbf{x}}_{t,b}\big\rVert_2$ \\[4pt]
$E_{\text{mpjve}}\downarrow$ & mm/s & Mean per-joint velocity error (finite difference) &
\makecell[l]{$\frac{10^3}{(T-1)J}\sum_{t=1}^{T-1}\sum_j \big\lVert (\mathbf{p}_{t+1,j}-\mathbf{p}_{t,j})$\\[2pt] $\quad -(\hat{\mathbf{p}}_{t+1,j}-\hat{\mathbf{p}}_{t,j})\big\rVert_2$} \\[4pt]
$E_{\text{mpjae}}\downarrow$ & mm/s$^2$ & Mean per-joint acceleration error (second finite difference) &
\makecell[l]{$\frac{10^3}{(T-2)J}\sum_{t=1}^{T-2}\sum_j \big\lVert (\mathbf{p}_{t+2,j}-2\mathbf{p}_{t+1,j}+\mathbf{p}_{t,j})$\\[2pt] $\quad -(\hat{\mathbf{p}}_{t+2,j}-2\hat{\mathbf{p}}_{t+1,j}+\hat{\mathbf{p}}_{t,j})\big\rVert_2$} \\[4pt]
$E_{\text{pbve}}\downarrow$ & cm/s & Root (pelvis) velocity error &
\makecell[l]{$\frac{10^2}{T-1}\sum_{t=1}^{T-1}\big\lVert (\mathbf{p}_{t+1,r}-\mathbf{p}_{t,r})$\\[2pt] $\quad -(\hat{\mathbf{p}}_{t+1,r}-\hat{\mathbf{p}}_{t,r})\big\rVert_2$} \\[4pt]
$E_{\text{pbae}}\downarrow$ & cm/s$^2$ & Root (pelvis) acceleration error &
\makecell[l]{$\frac{10^2}{T-2}\sum_{t=1}^{T-2}\big\lVert (\mathbf{p}_{t+2,r}-2\mathbf{p}_{t+1,r}+\mathbf{p}_{t,r})$\\[2pt] $\quad -(\hat{\mathbf{p}}_{t+2,r}-2\hat{\mathbf{p}}_{t+1,r}+\hat{\mathbf{p}}_{t,r})\big\rVert_2$} \\
\bottomrule
\end{tabular}
\end{table}

\bmhead{Baselines}
Since no continual-learning baseline exists for humanoid whole-body control, we (i) adapt three strong multi-motion humanoid imitation methods---\textbf{KungfuBot2}~\cite{han2025kungfubot2}, \textbf{ASAP}~\cite{he2025asap}, and \textbf{HumanoidVerse}~\cite{cum2025humanoidverse}---to our unified task-incremental protocol (each method is fine-tuned sequentially on a single network as tasks are added, with CL metrics computed after each addition), and (ii) adopt two strong CL methods, \textbf{EWC}~\cite{kirkpatrick2017overcoming} and \textbf{PackNet}~\cite{mallya2018packnet}. All methods, including ours, train only on the current task's data without replay; ours selects a parent column via motion similarity, trains only its LoRA adapter, and activates the evaluated task's column at test time.

\subsection{Main Results}
\label{subsec:main_results}

\begin{figure}[!htb]
\centering
\includegraphics[width=.9\textwidth]{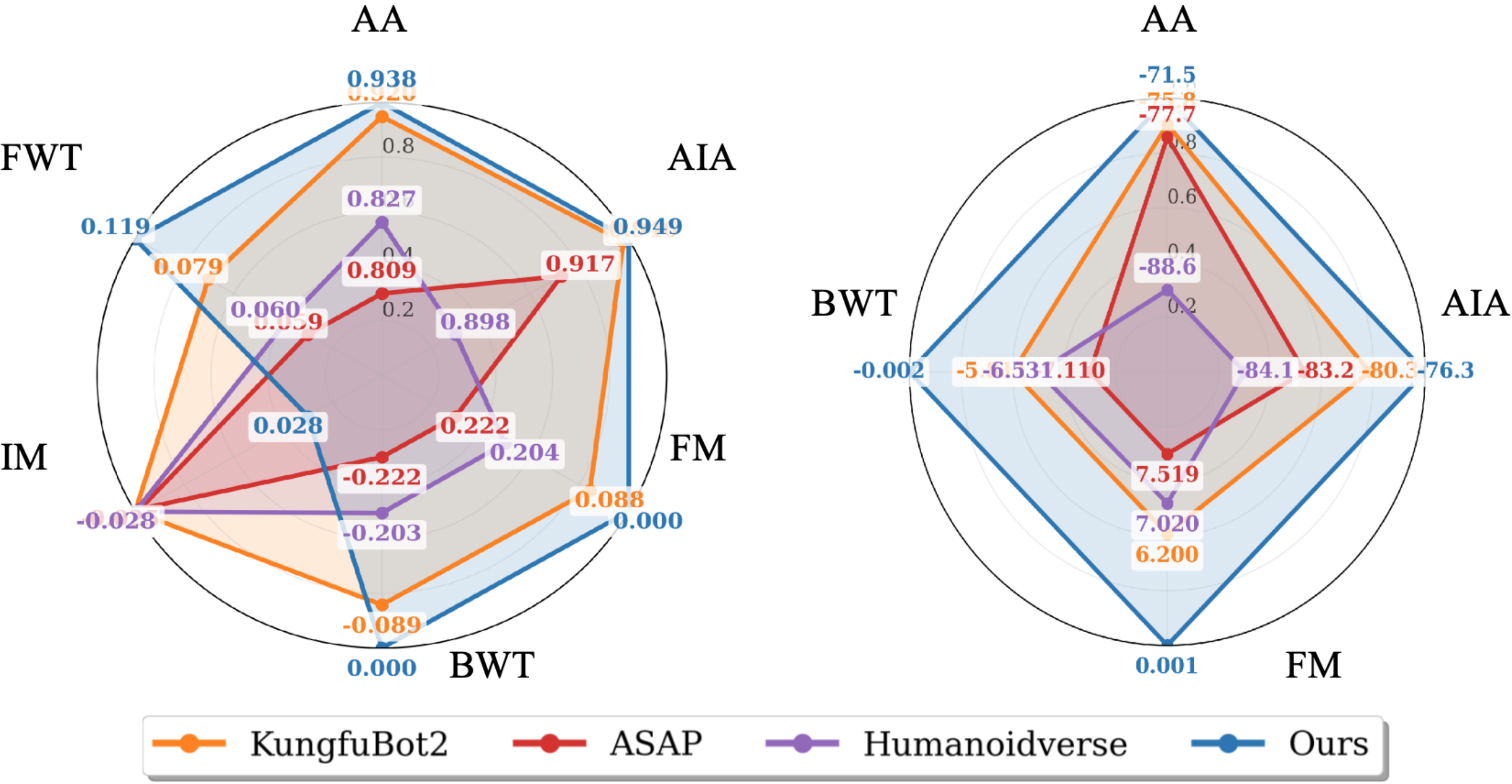}
\caption{Radar plots of \textbf{CL metrics.} (Left) Success rate. (Right) Negative MPJPE score. Polygons show min--max normalized values with square-root transformation for visual separation; numbers denote original metric values}
\label{fig:cl_radar}
\end{figure}

\bmhead{Continual-learning performance}
Table~\ref{tab:cl_main} and Fig.~\ref{fig:cl_radar} report CL metrics on success rate and $-$MPJPE. Our method achieves $\mathrm{FM}\!=\!\mathrm{BWT}\!\approx\!0$, keeping every previously learned skill intact. Our similarity-guided LoRA policy (\texttt{sim-LoRA}) attains the strongest forward transfer, $\mathrm{FWT}=0.125$ versus $0.079$ for the best baseline (KungfuBot2), together with the highest AA ($0.945$) and AIA ($0.964$) of all methods; the PNN full fine-tuning reference is close behind ($0.119/0.938/0.949$). All adapted baselines and CL methods exhibit measurable forgetting (FM $0.088$--$0.320$); EWC and PackNet, lacking motion-aware structure and lateral transfer, trail substantially in AA ($0.488$ and $0.586$). The one exception is intransigence: our IM ($0.021$) is slightly higher than that of the baselines (whose IM is negative), as frozen columns give up a small amount of plasticity relative to the joint-training reference in exchange for guaranteed retention. The radar plot in Fig.~\ref{fig:cl_radar} visualizes these results on both success rate and $-$MPJPE, confirming that improved continual success does not come at the cost of imitation accuracy.

\begin{table}[!htb]
\centering
\setlength{\tabcolsep}{5pt}
\caption{\textbf{Continual-learning metrics on success rate.} We compare adapted multi-motion humanoid baselines and CL methods under a unified task-incremental protocol. Results are mean (upper) and standard deviation (lower, $\pm$) over 100 evaluation episodes. IM/FWT require joint-training and single-task references that are not applicable to EWC/PackNet (marked \texttt{--}). Best in bold}
\label{tab:cl_main}
\begin{tabular}{lcccccc}
\toprule
\textbf{Method} & AA$\uparrow$ & AIA$\uparrow$ & FM$\downarrow$ & BWT$\uparrow$ & IM$\downarrow$ & FWT$\uparrow$ \\
\midrule
\multicolumn{7}{l}{\textit{Multi-motion humanoid baselines}}\\
KungfuBot2     & \ms{0.920}{0.058} & \ms{0.946}{0.053} & \ms{0.088}{0.006} & \ms{-0.089}{0.006} & \textbf{\ms{-0.028}{0.003}} & \ms{0.079}{0.005} \\
ASAP           & \ms{0.809}{0.044} & \ms{0.917}{0.052} & \ms{0.222}{0.122} & \ms{-0.222}{0.122} & \ms{-0.025}{0.003} & \ms{0.059}{0.003} \\
HumanoidVerse  & \ms{0.827}{0.023} & \ms{0.898}{0.026} & \ms{0.204}{0.007} & \ms{-0.203}{0.006} & \ms{-0.028}{0.000} & \ms{0.060}{0.002} \\
\midrule
\multicolumn{7}{l}{\textit{Standard CL methods}}\\
EWC            & \ms{0.488}{0.026} & \ms{0.645}{0.032} & \ms{0.220}{0.121} & \ms{-0.220}{0.120} & -- & -- \\
PackNet        & \ms{0.586}{0.032} & \ms{0.747}{0.037} & \ms{0.320}{0.176} & \ms{-0.320}{0.176} & -- & -- \\
\midrule
Ours (PNN, full FT)  & \ms{0.938}{0.043} & \ms{0.949}{0.044} & \ms{0.000}{0.000} & \ms{0.000}{0.000} & \ms{0.028}{0.001} & \ms{0.119}{0.006} \\
Ours (sim-LoRA)      & \textbf{\ms{0.945}{0.030}} & \textbf{\ms{0.964}{0.032}} & \textbf{\ms{0.000}{0.000}} & \textbf{\ms{0.000}{0.000}} & \ms{0.021}{0.001} & \textbf{\ms{0.125}{0.004}} \\
\bottomrule
\end{tabular}
\end{table}

\bmhead{Imitation quality}
Table~\ref{tab:mimic_metrics} confirms that the gains above do not degrade tracking: our policy retains \textbf{100\%} success and the \emph{lowest} dynamic errors ($E_{\text{mpjve}}$, $E_{\text{mpjae}}$), with pose errors comparable to strong single-task baselines. Table~\ref{tab:mimic_metrics} reports the \texttt{PNN (full FT)} configuration; in Table~\ref{tab:cl_main} we additionally report our similarity-guided LoRA variant (\texttt{sim-LoRA}, the \texttt{linear\_rank} setting), which is best on AA/AIA/FWT among all methods.

\begin{table}[!htb]
\centering
\setlength{\tabcolsep}{2.5pt}
\caption{\textbf{Imitation metrics} (single-task training, not CL). Success rate ($\uparrow$) and tracking errors ($\downarrow$); $E_{\text{mpjpe}}/E_{\text{gmpjpe}}/E_{\text{mpbpe}}$ in \si{mm}, $E_{\text{mpjve}}/E_{\text{mpjae}}$ in \si{mm/s} and \si{mm/s^2}, $E_{\text{pbve}}/E_{\text{pbae}}$ in \si{cm/s} and \si{cm/s^2}. Entries evaluated on our suite where available; \texttt{--} not applicable. Best in bold}
\label{tab:mimic_metrics}
\begin{tabular}{l c ccccccc}
\toprule
\textbf{Method}
& \textbf{Succ.}$\uparrow$
& $E_{\text{mpjpe}}\downarrow$
& $E_{\text{gmpjpe}}\downarrow$
& $E_{\text{mpbpe}}\downarrow$
& $E_{\text{mpjve}}\downarrow$
& $E_{\text{mpjae}}\downarrow$
& $E_{\text{pbve}}\downarrow$
& $E_{\text{pbae}}\downarrow$ \\
\midrule
KungfuBot2     & 100\%   & 87.38 & 147.84 & 29.81 & 105.64 & 90.45 & 4.27 & \textbf{3.54} \\
ASAP           & 100\%   & \textbf{77.0} & \textbf{129.0} & -- & -- & -- & \textbf{2.69} & 5.65 \\
HumanoidVerse  & 97.8\%  & 122.12 & 146.16 & 28.87 & 98.44 & 90.61 & 4.17 & 3.55 \\
Ours           & \textbf{100\%} & 80.7 & 135.36 & 30.67 & \textbf{91.79} & \textbf{84.92} & 4.72 & 4.23 \\
\bottomrule
\end{tabular}
\end{table}

\begin{table}[!htb]
\centering
\setlength{\tabcolsep}{3pt}
\caption{\textbf{Ablation and efficiency} after learning all $K$ tasks (CL metrics on success rate). We jointly ablate the architecture and report trainable parameters, training time per 5k iterations, and savings relative to PNN with full fine-tuning. FM/BWT are near-zero for all PNN-based variants; only the non-PNN ablation forgets. Best in bold}
\label{tab:efficiency_pnn}
\begin{tabular}{lccc|cccc}
\toprule
\textbf{Variant} & \textbf{AA}$\uparrow$ & \textbf{AIA}$\uparrow$ & \textbf{FM}$\downarrow$ &
\hd{\textbf{Trainable}\\\textbf{Params}$\downarrow$} & \hd{\textbf{Saved}\\\textbf{Params}\\(\%)$\uparrow$} &
\hd{\textbf{Time}\\(h/5k)$\downarrow$} & \hd{\textbf{Saved}\\\textbf{Time}\\(\%)$\uparrow$} \\
\midrule
w/o PNN (single backbone)   & \ms{0.843}{0.024} & \ms{0.892}{0.028} & \ms{0.137}{0.004} & 2.4M & -- & -- & -- \\
\midrule
PNN, full FT (w/o LoRA)     & \ms{0.938}{0.043} & \ms{0.949}{0.044} & \textbf{0.000} & 2.4M & -- & 2.94 & -- \\
PNN + LoRA, w/o sim.\ init  & \ms{0.896}{0.041} & \ms{0.923}{0.044} & \textbf{0.000} & \textbf{132K} & \textbf{94.50} & \textbf{1.74} & \textbf{40.82} \\
PNN + LoRA, fixed rank      & \ms{0.895}{0.038} & \ms{0.929}{0.041} & \textbf{0.000} & \textbf{132K} & \textbf{94.50} & \textbf{1.74} & \textbf{40.82} \\
PNN + LoRA, binned rank     & \ms{0.928}{0.025} & \ms{0.946}{0.036} & \textbf{0.000} & 182K & 92.42 & 2.28 & 22.55 \\
PNN + LoRA, linear rank     & \textbf{\ms{0.945}{0.030}} & \textbf{\ms{0.964}{0.032}} & \textbf{0.000} & 218K & 90.92 & 2.75 & 6.53 \\
\bottomrule
\end{tabular}
\end{table}

\subsection{Ablation Study}
\label{subsec:ablation}
Table~\ref{tab:efficiency_pnn} ablates each design choice. Removing PNN (a single shared backbone) collapses retention (FM $0.000\!\rightarrow\!0.137$, AA $0.945\!\rightarrow\!0.843$), confirming column isolation as the source of forgetting-free learning. Accuracy improves monotonically as rank allocation becomes similarity-aware (\texttt{fixed} $0.895$ $\rightarrow$ \texttt{binned} $0.928$ $\rightarrow$ \texttt{linear} $0.945$): the more closely the rank follows motion similarity, the better each column's capacity matches its task, and \texttt{linear\_rank} attains the best AA/AIA ($0.945/0.964$). Similarity-guided initialization acts mainly on convergence and forward transfer, consistent with the higher FWT in Table~\ref{tab:cl_main}.

\begin{figure}[!htb]
  \centering
  \includegraphics[width=.7\textwidth]{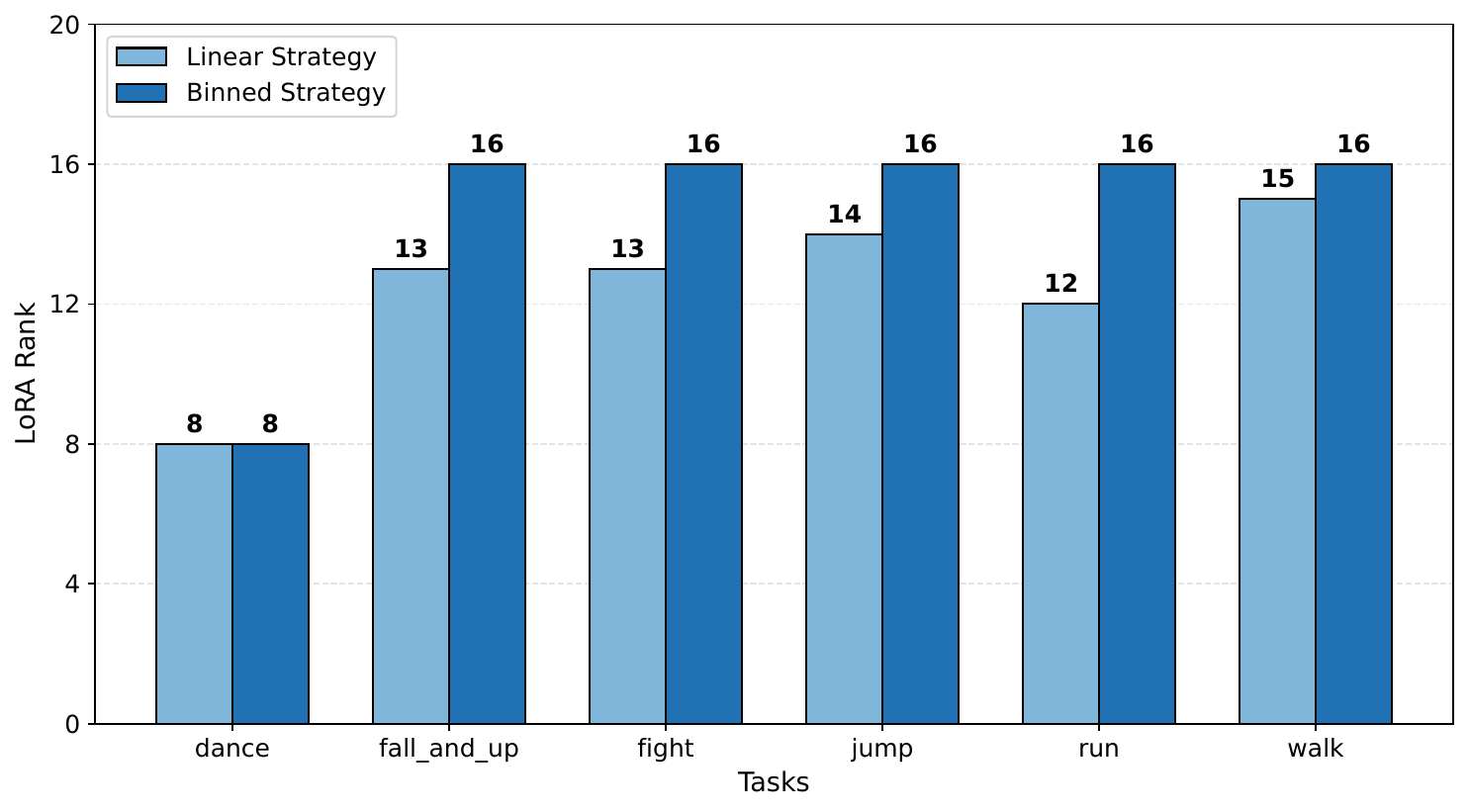}
  \caption{\textbf{Similarity-guided LoRA rank allocation.} Rank assignment under the \textbf{linear} rule, which smoothly increases rank as task similarity decreases, and the \textbf{binned} rule, which maps similarity into discrete rank levels}
  \label{fig:lora_rank_alloc}
\end{figure}

\bmhead{Rank allocation}
We allocate LoRA ranks in a similarity-guided manner, assigning larger ranks to tasks that are less similar to those learned previously (Table~\ref{tab:lora_rule}). Fig.~\ref{fig:lora_rank_alloc} visualizes the resulting assignment under the linear and binned rules. This concentrates adaptation capacity where knowledge transfer is likely to be limited, while preserving parameter efficiency for tasks with stronger similarity and higher expected transfer.

Table~\ref{tab:lora_rank_cl_metrics} complements the success-rate ablation with the same comparison on $-$MPJPE. Similarity-guided rank allocation consistently improves over the fixed-rank baseline: on success rate the linear rule achieves the best AA and AIA, while the binned rule provides smaller but clear gains. The same trend holds for motion quality: the linear rule yields the highest $-$MPJPE (\ie, the lowest MPJPE) for both AA and AIA, and the binned rule remains superior to fixed rank, indicating that allocating more capacity to less transferable tasks improves both task success and imitation fidelity.

\begin{table}[!htb]
\centering
\caption{\textbf{CL metrics under different LoRA rank-allocation rules,} on success rate and on $-$MPJPE (higher is better for both). Mean (upper) and standard deviation (lower, $\pm$) over 100 evaluation episodes per task. FM and BWT are omitted because all PNN-based variants exhibit near-zero forgetting. Best in bold}
\label{tab:lora_rank_cl_metrics}
\setlength{\tabcolsep}{8pt}
\begin{tabular}{lcccc}
\toprule
 & \multicolumn{2}{c}{\textbf{Success rate}} & \multicolumn{2}{c}{\textbf{$-$MPJPE}} \\
\cmidrule(lr){2-3}\cmidrule(lr){4-5}
\textbf{Rank rule} & AA$\uparrow$ & AIA$\uparrow$ & AA$\uparrow$ & AIA$\uparrow$ \\
\midrule
Fixed  & \ms{0.895}{0.038} & \ms{0.929}{0.041} & \ms{-116.542}{3.793} & \ms{-128.423}{4.841} \\
Binned & \ms{0.928}{0.025} & \ms{0.946}{0.036} & \ms{-98.062}{2.783} & \ms{-94.542}{3.767} \\
Linear & \textbf{\ms{0.945}{0.030}} & \textbf{\ms{0.964}{0.032}} & \textbf{\ms{-88.524}{4.173}} & \textbf{\ms{-94.303}{2.510}} \\
\bottomrule
\end{tabular}
\end{table}

\subsection{Efficiency}
\label{subsec:efficiency}
Beyond accuracy, similarity-guided LoRA makes progressive expansion efficient (Table~\ref{tab:efficiency_pnn}). Replacing full fine-tuning with LoRA cuts trainable parameters by up to $94.5\%$ ($2.4\mathrm{M}\!\rightarrow\!132\mathrm{K}$) and training time by $40.8\%$ ($2.94\!\rightarrow\!1.74$~h per 5k iterations), while even the best-accuracy \texttt{linear\_rank} variant still saves $90.9\%$ of trainable parameters. The four variants thus trace an accuracy--efficiency trade-off in which the cost of a new skill scales with its novelty rather than a fixed per-column budget: a skill close to an already-learned one is acquired with a small low-rank update, whereas a dissimilar skill receives a larger rank, so new parameters are spent where transfer is weakest. These savings also bring scalability: although PNN's lateral connections grow as $O(K^2)$ in the number of tasks, similarity-guided inheritance, LoRA, and parallel training of disjoint columns keep the per-task cost low and make substantially longer task streams practical.

\begin{figure}[t]
\centering
\includegraphics[width=\textwidth]{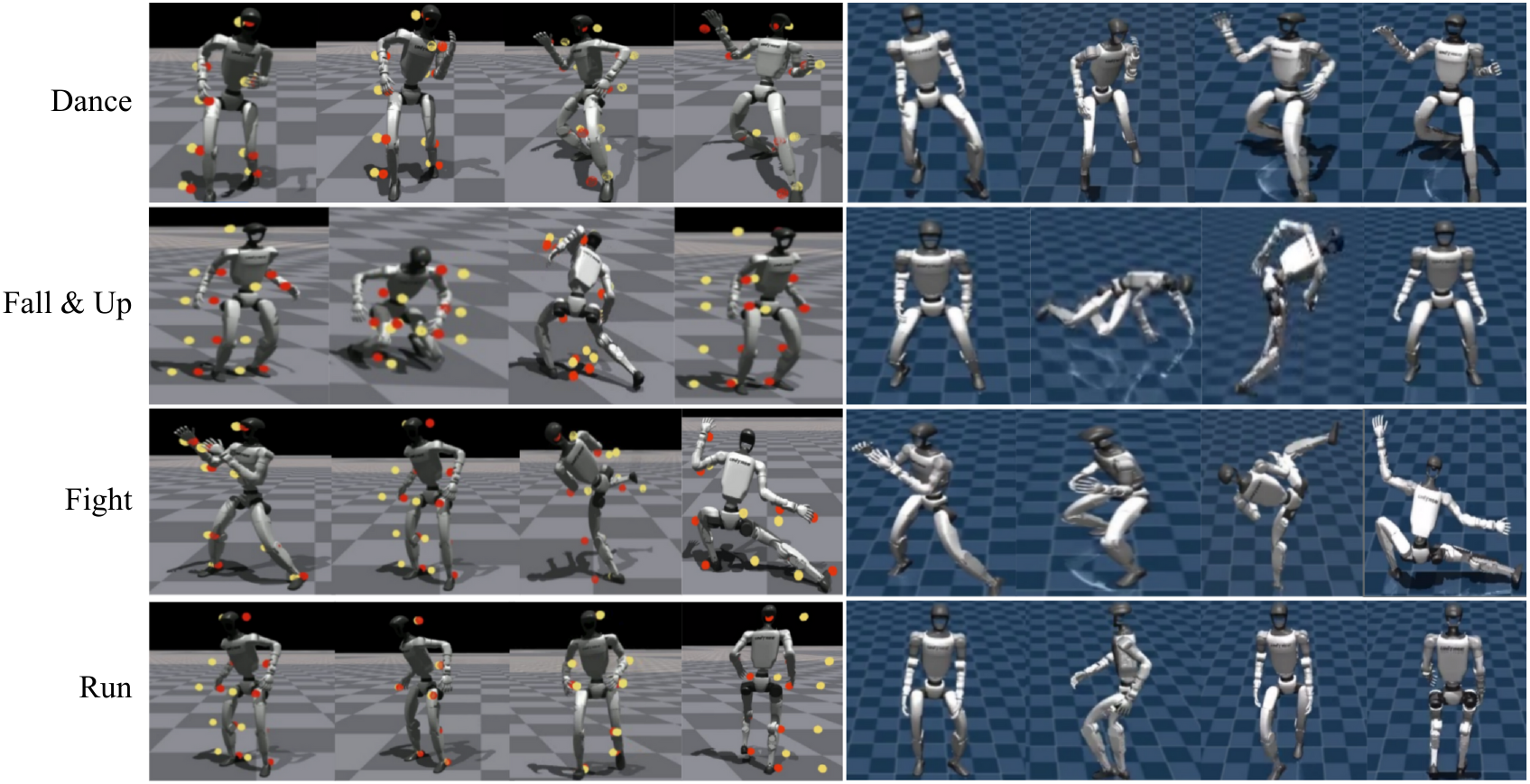}
\caption{Qualitative sim-to-sim comparison of the single distilled student on \texttt{dance}, \texttt{fall\_and\_up}, \texttt{fight}, and \texttt{run}: the same policy executed in Isaac Gym (left, with reference key-body markers) and MuJoCo (right)}
\label{fig:sim2sim_vis}
\end{figure}

\subsection{Real-World Deployment}
\label{subsec:deployment}
We validate deployment through a two-stage transfer pipeline: \emph{sim-to-sim} from Isaac Gym to MuJoCo with the distilled student, then the same student on a physical Unitree~G1 ($23$ actuated DoF; wrists locked for safety). The student attains a $96.13\%$ sim-to-sim success rate ($50$ episodes per motion, $1500$ trials) with only modest tracking degradation (Table~\ref{tab:sim2sim}); Fig.~\ref{fig:sim2sim_vis} shows the same motions executed consistently in both simulators.

\bmhead{Continual distillation}
Distilling a \emph{continual} teacher into a \emph{continual} student preserves transfer: it reaches $96.13\%$ sim-to-sim success, versus $94.07\%$ when task-specific teachers are merged into one student (Table~\ref{tab:teacher_student}), confirming that the progressive policy distills into a single compact controller without losing fidelity.

\begin{figure}[!htb]
\centering
\includegraphics[width=\textwidth]{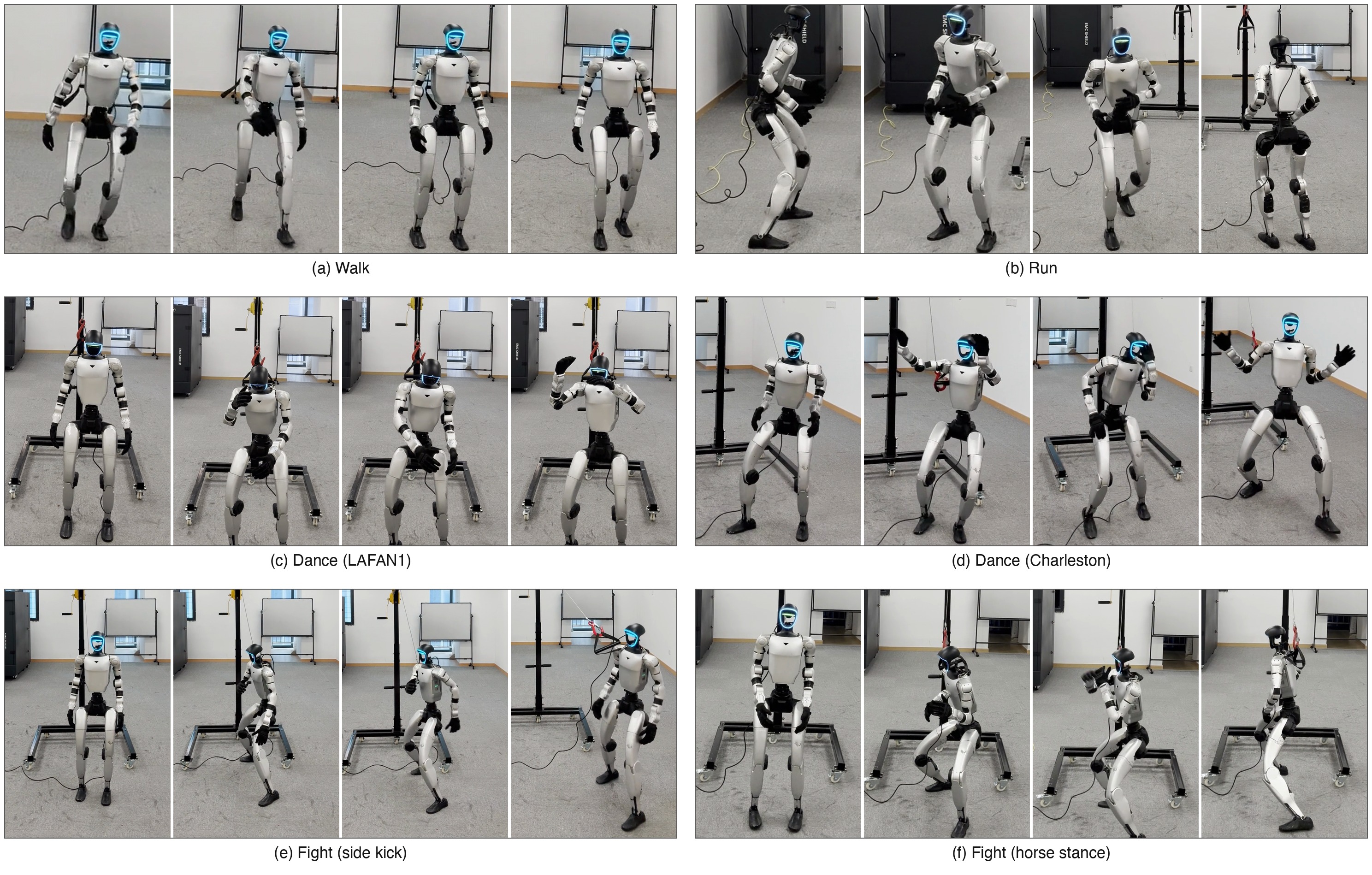}
\caption{\textbf{Real-robot execution} of the single continually trained student on the Unitree~G1. Each panel shows one motion, with frames ordered left to right in time: (a) walk, (b) run, (c)--(d) two dance motions, and (e)--(f) two fight motions (a side kick and a horse-stance sequence). All six motions are executed by the same controller without any per-skill fine-tuning}
\label{fig:real_robot}
\end{figure}

\bmhead{Real-robot results}
We deploy the \emph{single} continually trained student on the Unitree~G1 across $30$ motions spanning \emph{all six} skill categories, attaining $90.33\%$ success over $10$ episodes per motion under the same criterion. Fig.~\ref{fig:real_robot} shows representative frames of the deployed controller executing walking, running, dancing, and fighting motions on the physical robot; the full executions are in the supplementary video (Supplementary Video~1). To our knowledge this is among the first continually learned humanoid whole-body controllers deployed on hardware across its sequentially acquired skill set.

\begin{table}[t]
\centering
\caption{\textbf{Sim-to-sim transfer (Isaac Gym $\rightarrow$ MuJoCo).} Transfer success rate of the deployable student, and relative change ($\Delta$, \%) in MPJPE/MPJVE after transfer}
\label{tab:sim2sim}
\setlength{\tabcolsep}{4pt}
\begin{tabular}{ccc}
\toprule
Transfer Success Rate (\%) & $\Delta E_{\text{mpjpe}}$ (\%) & $\Delta E_{\text{mpjve}}$ (\%) \\
\midrule
$96.13$ & $+6.52$ & $+14.1$ \\
\bottomrule
\end{tabular}
\end{table}
\begin{table}[t]
\centering
\caption{\textbf{Continual distillation ablation.} Sim-to-sim success rate of the deployable student under two distillation strategies. Our continual teacher $\rightarrow$ continual student yields a single deployable controller while retaining all skills}
\label{tab:teacher_student}
\setlength{\tabcolsep}{4pt}
\begin{tabular}{lc}
\toprule
Distillation strategy & Sim-to-sim SR (\%) \\
\midrule
Task-wise teachers $\rightarrow$ unified student & $94.07$ \\
Continual teacher $\rightarrow$ continual student (\textbf{Ours}) & $\mathbf{96.13}$ \\
\bottomrule
\end{tabular}
\end{table}

\section{Conclusion}
\label{sec:conclusion}
We introduced a continual-learning framework for humanoid whole-body motion that lets a single controller sequentially acquire diverse skills without forgetting previously learned ones, combining a progressive policy with LoRA updates guided by motion similarity. It delivers the strongest forward transfer and large efficiency gains---up to $94.5\%$ fewer trainable parameters and $40.8\%$ less training time---and transfers to a physical Unitree~G1.

\backmatter

\setcounter{mainfig}{\value{figure}}%
\setcounter{maintab}{\value{table}}%
\begin{appendices}
\setcounter{figure}{\value{mainfig}}%
\setcounter{table}{\value{maintab}}%
\renewcommand{\thefigure}{\arabic{figure}}%
\renewcommand{\thetable}{\arabic{table}}%

\section{Motion Data Format and Implementation Notes}
\label{app:sim}
This appendix complements Sec.~\ref{subsec:sim} with the data format and the implementation constants of the motion-set distance.

\bmhead{Motion data format}
Each retargeted motion carries the following attributes:
\begin{itemize}
\item \textbf{Root translation}: \texttt{root\_trans\_offset} $\in \mathbb{R}^{T\times 3}$, the root position over time in a fixed world frame.
\item \textbf{Root orientation}: \texttt{root\_rot} $\in \mathbb{R}^{T\times 4}$, the quaternion root rotation over time in a fixed world frame.
\item \textbf{Pose parameters}: \texttt{pose\_aa} $\in \mathbb{R}^{T\times J\times 3}$, axis-angle rotations for $J$ joints.
\item \textbf{Optional contact}: \texttt{contact\_mask} $\in \mathbb{R}^{T\times K}$, per-frame contact signals whose values may be $\{0,1\}$ or soft values (\eg, $0.5$).
\item \texttt{fps}: sampling frequency of the sequence.
\end{itemize}
Here $T$ is the motion length, which varies across motions, and $K$ is the number of contact points of the humanoid on the ground. The planar root position $p_t$ and yaw $\psi_t$ used in Eq.~\ref{eq:features} are read from \texttt{root\_trans\_offset} and \texttt{root\_rot}; the joint rotations $R_{t,j}$ are obtained from \texttt{pose\_aa}. The per-frame feature $f_t$ is defined for $t\in\{0,\dots,T-2\}$ because every term is a forward difference.

\bmhead{Constants}
All motions are resampled to $f_0=\SI{30}{Hz}$. The Sakoe--Chiba band half-width is $b=\lfloor\rho L_f\rfloor$ with $\rho\in[0.10,0.15]$. The four window/stride configurations are $(L,S)\in\{(3,2),(5,3),(10,7),(15,10)\}$\,s (Table~\ref{tab:hparams}). The feature weights $w_v,w_\omega,w_J,w_c$ in Eq.~\ref{eq:frame_cost} and the temporal-regularization weight $\beta$ in Eq.~\ref{eq:dmotion} are fixed and shared across all task pairs.

\bmhead{Remarks}
The $\ell_1$ term in Eq.~\ref{eq:frame_cost} is used for contacts because contact labels may be soft real values. Path-length normalization in Eq.~\ref{eq:window_cost} reduces sensitivity to minor differences in DTW path length. Since both OT stages use uniform masses, a difference between $D_{\text{set}}(\mathcal{A},\mathcal{B})$ and $D_{\text{set}}(\mathcal{C},\mathcal{D})$ arises from motion patterns and their coverage rather than from $N+M$ versus $N^\prime+M^\prime$; in practice, larger sets yield more stable estimates due to reduced sampling variance.

\section{Hyperparameters and Implementation Details}
\label{app:hparams}
This appendix expands the summary in Table~\ref{tab:hparams}. Unless otherwise specified, a single hyperparameter setting is shared across all tasks. The motion-distance settings are given in Appendix~\ref{app:sim} and Table~\ref{tab:hparams}.

\bmhead{PPO training}
All policies are trained with PPO~\cite{schulman2017proximal} under a shared configuration: fixed rollout horizon, mini-batch count, and learning rate, with multiple PPO epochs per update, generalized advantage estimation, and the standard clipped policy and value objectives (Table~\ref{tab:ppo_hparams}).

\begin{table}[!htb]
\centering
\caption{\textbf{PPO training hyperparameters}}
\label{tab:ppo_hparams}
\setlength{\tabcolsep}{5pt}
\begin{tabular}{l c >{\raggedright\arraybackslash}p{0.45\textwidth}}
\toprule
\textbf{Parameter} & \textbf{Value} & \textbf{Description} \\
\midrule
Learning rate & $1\mathrm{e}{-4}$ & For both actor and critic \\
Clip parameter $\epsilon$ & $0.2$ & PPO clipping range \\
Discount factor $\gamma$ & $0.99$ & Reward discount \\
GAE $\lambda$ & $0.95$ & Generalized advantage estimation \\
Value-loss coefficient & $1.0$ & Weight of the value-function loss \\
Entropy coefficient & $0.01$ & Entropy regularization weight \\
Max gradient norm & $1.0$ & Gradient clipping threshold \\
Learning epochs & $5$ & PPO epochs per update \\
Mini-batches & $4$ & Mini-batches per update \\
Steps per environment & $24$ & Collected steps per environment per rollout \\
KL target & $0.01$ & Target KL for the adaptive learning rate \\
Initial action-noise std & $1.0$ & Initial standard deviation of the action noise \\
Training iterations & $50{,}000$ & Per task \\
Parallel environments & $1024$ & Isaac Gym \\
\bottomrule
\end{tabular}
\end{table}

\bmhead{PNN architecture}
The continual-learning backbone follows the Progressive Neural Network paradigm~\cite{rusu2016progressive}: a new column is instantiated for each task while all previously learned columns are frozen, the current column receives lateral connections from earlier columns, and (when enabled) LoRA adapters are optimized only for the current task (Table~\ref{tab:pnn_arch}).

\begin{table}[!htb]
\centering
\caption{\textbf{PNN network architecture}}
\label{tab:pnn_arch}
\setlength{\tabcolsep}{5pt}
\begin{tabular}{l >{\raggedright\arraybackslash}p{0.62\textwidth}}
\toprule
\textbf{Component} & \textbf{Configuration} \\
\midrule
Motion encoder & Conv1D, hidden dimension $60$, output dimension $128$, SiLU \\
History encoder & Conv1D, hidden dimension $30$, output dimension $64$, SiLU \\
Privileged encoder & MLP $[64]$, output dimension $64$, SiLU \\
Task column (actor) & MLP $[512,\,512,\,256]$, SiLU \\
Critic & MLP $[768,\,512,\,256]$, SiLU, LayerNorm \\
Lateral connection & Linear adapter, zero-initialized \\
Max tasks & $16$ \\
\bottomrule
\end{tabular}
\end{table}

\bmhead{Domain randomization}
Randomized physical, actuation, and control parameters are sampled at episode reset and held constant throughout the episode (Table~\ref{tab:domain_randomization}).

\begin{table}[!htb]
\centering
\caption{\textbf{Domain randomization settings}}
\label{tab:domain_randomization}
\setlength{\tabcolsep}{8pt}
\begin{tabular}{l >{\raggedright\arraybackslash}p{0.55\textwidth}}
\toprule
\textbf{Parameter} & \textbf{Range} \\
\midrule
Link mass & Scaled by $[0.9,\,1.1]$ \\
Link inertia & Scaled by $[0.9,\,1.1]$ \\
Base COM offset & $x,y$: $\pm\SI{0.05}{m}$; $z$: $\pm\SI{0.01}{m}$ \\
PD gains ($K_p$, $K_d$) & Scaled by $[0.9,\,1.1]$ \\
Ground friction & Uniform in $[0.2,\,1.2]$ \\
Torque RFI & Additive noise in $\pm0.05$ \\
RAO & Additive noise in $\pm0.05$ \\
Control delay & Discrete delay in $[0,\,2]$ steps ($0$--$\SI{40}{ms}$) \\
Push velocity & Maximum magnitude $\SI{0.1}{m/s}$ \\
Push interval & Uniform in $[5,\,10]\,\si{s}$ \\
\bottomrule
\end{tabular}
\end{table}

\bmhead{Reward and termination}
The reward comprises motion-tracking terms and shared regularization penalties (Eq.~\ref{eq:reward_sum}); episodes terminate early on severe instability or safety violations. All weights and coefficients are listed in Table~\ref{tab:reward_termination}.

\begin{table}[!htb]
\centering
\caption{\textbf{Reward and termination design.} Tracking rewards use $r = w\exp(-\lVert e\rVert^2/\sigma^2)$, where $e$ is the tracking error}
\label{tab:reward_termination}
\setlength{\tabcolsep}{5pt}
\begin{tabular}{l c c >{\raggedright\arraybackslash}p{0.30\textwidth}}
\toprule
\multicolumn{4}{l}{\textit{Motion-tracking rewards}} \\
\textbf{Term} & \textbf{Weight $w$} & $\boldsymbol{\sigma}$ & \textbf{Description} \\
\midrule
\texttt{teleop\us root\us vel} & 1.0 & 1.0 & Root velocity tracking \\
\texttt{teleop\us anchor\us body\us position} & 1.5 & 0.1 & Anchor-body position tracking \\
\texttt{teleop\us anchor\us body\us rotation} & 1.5 & 0.15 & Anchor-body rotation tracking \\
\texttt{local\us key\us body\us position} & 3.0 & 0.1 & Local key-body position tracking \\
\texttt{local\us key\us body\us rotation} & 2.0 & 0.15 & Local key-body rotation tracking \\
\texttt{key\us body\us velocity} & 2.0 & 1.0 & Key-body linear-velocity tracking \\
\texttt{key\us body\us ang\us velocity} & 2.0 & 10.0 & Key-body angular-velocity tracking \\
\midrule
\multicolumn{4}{l}{\textit{Penalty and termination terms}} \\
\textbf{Term} & \textbf{Weight} & \multicolumn{2}{l}{\textbf{Description}} \\
\midrule
\texttt{penalty\us action\us rate} & $-0.1$ & \multicolumn{2}{>{\raggedright\arraybackslash}p{0.36\textwidth}}{Rapid changes in actions} \\
\texttt{penalty\us dof\us vel} & $-1.0\times10^{-4}$ & \multicolumn{2}{>{\raggedright\arraybackslash}p{0.36\textwidth}}{Large joint velocities} \\
\texttt{penalty\us dof\us acc} & $-3.0\times10^{-7}$ & \multicolumn{2}{>{\raggedright\arraybackslash}p{0.36\textwidth}}{Large joint accelerations} \\
\texttt{foot\us slip\us penalty} & $-0.1$ & \multicolumn{2}{>{\raggedright\arraybackslash}p{0.36\textwidth}}{Foot slipping during contact} \\
\texttt{limits\us dof\us pos} & $-10.0$ & \multicolumn{2}{>{\raggedright\arraybackslash}p{0.36\textwidth}}{Joint-position limit violations} \\
\texttt{limits\us torque} & $-5.0$ & \multicolumn{2}{>{\raggedright\arraybackslash}p{0.36\textwidth}}{Actuator torque limit violations} \\
\texttt{collision} & $-30.0$ & \multicolumn{2}{>{\raggedright\arraybackslash}p{0.36\textwidth}}{Body collisions} \\
\texttt{termination} & $-200.0$ & \multicolumn{2}{>{\raggedright\arraybackslash}p{0.36\textwidth}}{Applied upon episode termination} \\
\bottomrule
\end{tabular}
\end{table}
\end{appendices}
\clearpage   %

\bibliography{references}

\section*{Statements and Declarations}

\bmhead{Competing Interests}
The authors have no competing interests to declare that are relevant to the
content of this article.

\bmhead{Author Contributions}
Zhewen He designed the method, implemented the code, conducted the simulation
experiments, took part in the real-robot deployment, and wrote the first draft
of the manuscript. Hao Huang advised on the design of the method and carried
out the real-robot deployment together with Zhewen He and Geeta Chandra Raju
Bethala. Anthony Tzes and Yi Fang provided guidance on the research and
secured the funding. All authors commented on previous
versions of the manuscript and approved the final version.

\bmhead{Data Availability}
The code and configuration files used in this study are available at
\url{https://anonymous.4open.science/r/continual-humanoid-learning-35D3}.
The LAFAN1 and Kungfu motion datasets are publicly available from their
original authors~\cite{harvey2020robust,xie2025kungfubot,han2025kungfubot2}.

\bmhead{Supplementary Information}
Supplementary Video~1 (provided with this preprint as an ancillary file; \texttt{supplementary\_video.mp4}, MP4/H.264, 1920$\times$1080, 3~min
50~s, with English narration): supplementary video showing (i) the
catastrophic-forgetting motivation, (ii) an overview of the similarity-guided
LoRA-PNN pipeline, (iii) the single continually trained controller on the
physical Unitree~G1 learning \texttt{walk}, \texttt{run}, \texttt{dance}, and
\texttt{fight} in sequence and then recalling these four categories in reverse order,
and (iv) sim-to-sim transfer and the comparison against the baselines.

\end{document}